# SIGNAL TO NOISE RATIO IN LENSLESS COMPRESSIVE IMAGING

Hong Jiang, Gang Huang and Paul Wilford

Bell Labs, Alcatel-Lucent, Murray Hill, NJ 07974

*Abstract*—We analyze the signal to noise ratio (SNR) in a lensless compressive imaging (LCI) architecture. The architecture consists of a sensor of a single detecting element and an aperture assembly of an array of programmable elements. LCI can be used in conjunction with compressive sensing to capture images in a compressed form of compressive measurements. In this paper, we perform SNR analysis of the LCI and compare it with imaging with a pinhole or a lens. We will show that the SNR in the LCI is independent of the image resolution, while the SNR in either pinhole aperture imaging or lens aperture imaging decreases as the image resolution increases. Consequently, the SNR in the LCI is much higher if the image resolution is large enough.

## I. INTRODUCTION

LENSLESS compressive imaging (LCI) [1] is an effective architecture to acquire images using the compressive sensing technique [2][3]. It consists of a sensor of a single detector element and an aperture assembly, but no lens is used, as illustrated in Figure 1. The transmittance of each aperture element is individually programmable. The sensor can be used to acquire compressive measurements which, in turn, can be used to reconstruct an image of the scene. By using compressive sensing, an image can be reconstructed using far fewer measurements than the number of pixels in the image, and therefore, an image is already compressed when it is acquired in the form of compressive measurements. This architecture is distinctive in that the images acquired are not formed by any physical mechanism, such as a lens [4] or a pinhole [5]-[7], and therefore, there are no aberrations introduced by a lens, such as a scene being out of focus. Furthermore, the same architecture can be used for acquiring multimodal signals such as infrared, Terahertz [8] and millimeter wave images [9]. This architecture has application in surveillance [10].

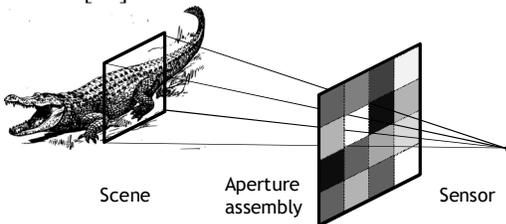

**Figure 1.** Lensless compressive imaging (LCI) architecture

Since the LCI architecture of Figure 1 does not use a lens, does it suffer from poor signal to noise ratio (SNR)? How does its SNR compare to that of a pinhole aperture imaging (PAI) or to that of a digital camera with a lens, i.e., the lens aperture imaging (LAI)? The goal of this paper is to answer these questions. We will perform SNR analysis and compare LCI with the pinhole aperture imaging (LCI) and the lens aperture imaging (LAI).

The LCI architecture allows an image to be acquired directly as compressed data, in the form of compressive measurements. There are two types of noises in the final reconstructed image: measurement noise and compression noise. The measurement noise is defined as the noise present in the process of acquiring data from the imaging device, such as shot noise, thermal noise and quantization noise in the acquired data. The compression noise is defined as noise due to compression, i.e., the error introduced in the reconstruction because not all compressive measurements are used, even if the measurements themselves are acquired precisely, free of any measurement noise.

Similarly, in addition to the measurement noise, the PAI or LAI may also be subject to compression noise, which is the error introduced when an image is compressed, e.g., by using the JPEG compression.

In this paper, we only analyze the measurement noise. The analysis of compression noise in LCI, i.e., errors due to the use of partial measurements in reconstruction, can be found in the compressive sensing literature, e.g., [2][3]. In neglecting the compression noise, we assume all measurements are used in reconstruction and analyze SNR of the reconstructed image due to the measurement noise. Two types of noises are included in the measurement noise. The first type is the shot noise, due to statistical quantum fluctuations, which is modeled by a Poisson distribution. The second type is additive noise which includes thermal noise, quantization noise etc., and is modeled by a random variable of zero mean and certain variance. There is no assumption on the type of distribution for additive noise.

Our analysis will show that the SNR of LCI can be better than that of PAI or LAI. More specifically, how the SNR of LCI compares with that of PAI or LAI depends on, among other parameters, the resolution of the image. Our analysis will show that, when other parameters, such as noise characteristics of sensors, the size of lens aperture etc, are fixed, the LCI will have a better SNR than PAI or LAI if the resolution, i.e., the number of pixels, of the image is large enough. This result is significant because it shows that even without a lens, the LCI can have a higher SNR than a digital camera with a lens.

## II. LENSLESS COMPRESSIVE IMAGING

The LCI architecture of Figure 1 can be used to create an image $x$ by using compressive sensing, see [1] for details. When the pixels of the image are arranged as a 1D vector, $x$ is a vector of length $N$ whose component $x_i$ is the light intensity from pixel $i$ of the object plane. Given a sensing matrix $A$, the measurement vector, in the absence of measurement noise, is given by

$$y = Ax. \qquad (1)$$

As noted before, we will only consider measurement noise, the noise that is introduced into the measurements while they are acquired. The compression noise, the error due to reconstruction by using only portion of measurements, is not considered in this paper. Therefore, we assume that the sensing matrix $A$ is a square matrix, i.e., there are as many measurements as the number of pixels in an image, and all measurements are used in the reconstruction of the image.

A randomly permutated Hadamard matrix is commonly used as a sensing matrix for image or video [1][10]-[12]. Since the random permutation does not change the noise analysis, we will omit permutation and use a modified Hadamard matrix as the sensing matrix, in which the entries of -1 in the Hadamard matrix are replaced by 0 because the transmittance in LCI is not defined for -1. Let $H = [h_{ij}]$ be $N \times N$ Hadamard matrix. Then the entries of our sensing matrix $A = [a_{ij}]$ are given by

$$a_{ij} = 1, \text{if } h_{ij} = 1, a_{ij} = 0, \text{if } h_{ij} = -1, i,j = 1,...,N. \quad (2)$$

*A. Measurement noise*

The measurement vector $y$ in (1) can be contaminated by measurement noise. We denote by $z$ the vector of acquired values of the measurement vector $y$ in the presence of measurement noise and will establish a relationship between $z$ and $y$ by making some assumptions on the noise. We consider two types of measurement noise.

**Shot Noise**:

The shot noise is caused by statistical quantum fluctuations in the number of photons collected by the sensor, and it is modeled by the Poisson distribution. Therefore, the actual acquired value from the sensor for measurement $y_i$ is a random variable $\hat{y}_i$ of Poisson distribution $P(y_i)$, i.e.,

$$\begin{aligned} \hat{y}_i &\sim P(y_i) \\ E(\hat{y}_i) &= \text{var}(\hat{y}_i) = y_i \end{aligned} \quad i = 1,...,N. \quad (3)$$

In (3), $E(\cdot)$ and $\text{var}(\cdot)$ denote the expected value and the variance, respectively. We further assume that $\hat{y}_i$ are independent random variables.

**Additive Noise**:

This type of noise has a fixed variance independent of light intensity. This noise includes thermal noise and quantization noise in the measurements, and is modeled by a random variable $\varepsilon_i$ for measurement $y_i$. We further assume that $\varepsilon_i$ are independent and identically distributed random variables with zero mean and a variance of $\sigma^2$, i.e.,

$$E(\varepsilon_i) = 0, \text{var}(\varepsilon_i) = \sigma^2, i = 1,...,N. \quad (4)$$

Under the assumption of above two types of noise, the actual acquired value for $y_i$, read from the sensor in the presence of the measurement noise, can be written as

$$z_i = \hat{y}_i + \varepsilon_i, i = 1,...,N, \text{ or } z = \hat{y} + \varepsilon. \quad (5)$$

The vector $z$ of (5) is the actual acquired data of the measurement vector $y$ from the sensor in the presence of measurement noise.

*B. Signal to noise ratio*

In LCI, the image $x$ is not acquired directly; instead, it must be reconstructed from the acquired measurement vector $z$. Reconstruction algorithms are well known in compressive sensing literature, see for example [1] [12], but in the context of this paper, since our sensing matrix is a square, invertible matrix, the reconstruction can be performed simply by solving (1) for $x$, with $y$ replaced by $z$.

Let $\tilde{x}$ be the image reconstructed from the acquired measurement vector $z$, i.e.,

$$\tilde{x} = A^{-1} z. \quad (6)$$

Then the goal is to find the SNR of the reconstructed image $\tilde{x}$.

We consider the total signal power and total noise power in the entire image $\tilde{x}$. The total signal power is the integration of all light rays from the scene to the sensor when all aperture elements are open, see Figure 1, and it is given by

$$X^0 = \sum_{i=1}^{N} x_i. \quad (7)$$

The value $X^0$ defined in (7) is the brightness of the scene as seen by the sensor, and it is only a function of the lighting of the scene and the field of view.

The SNR of LCI due to measurement noise is defined as the ratio of the total signal power in the image to the total noise power in the image, given by

$$\text{SNR}^{LCI} = X^0 \Big/ \sqrt{\sum_{i=1}^{N} \text{var}(\tilde{x}_i)}. \quad (8)$$

By computing the variance of $\tilde{x}$ from (6), we can show the following result.

**Proposition 1.**

*If the sensing matrix is the modified Hadamard matrix given in* (2), *then we have*

$$\text{SNR}^{LCI} = \frac{X^0}{\sqrt{\left(2 - \frac{4}{N}\right) X^0 + \left(4 - \frac{4}{N}\right)\sigma^2}} \geq \frac{X^0}{\sqrt{2X^0 + 4\sigma^2}} \quad (9)$$

*where $\sigma^2$ is the variance of the additive noise given in (4)*.

The lower bound for the SNR in (9) is only a function of the brightness $X^0$ and the power of the additive noise, $\sigma^2$. In the denominator, the value $\sqrt{X^0}$ represents the total power of shot noise, and $\sigma$ represents the power of additive noise when the sensor acquires each measurement.

An important observation from Proposition 1 is that the lower bound in (9) is independent of the image resolution $N$. The SNR of LCI due to measurement noise is bounded below by a constant with respect to the resolution, and in particular, it does not reduce when the image resolution $N$ increases.

### III. COMPARISON

In this section, we present SNR results for two other imaging architectures: the pinhole aperture imaging (PAI), and the lens aperture imaging (LAI). We assume that images in all architectures have the same resolution, i.e., the same number of pixels, $N$, which means that the number of aperture elements in LCI is the same as the number of sensors in PAI and LAI. We further assume that the scene and the field of view of the images are the same in all architectures to be compared, and the field of view does not change with the resolution $N$. A corollary of these assumptions is that the brightness of the scene, $X^0$ defined in (7), is independent of resolution $N$.

We will compare the SNR of LCI due to measurement noise with each of the PAI and LAI and show that the former outperforms both the PAI and LAI if the image resolution is high enough, i.e., if $N$ is large enough.

*A. Measurement noise*

Similar to the previous section, the two types of the noise are modeled in PAI and LAI. What is different in this section is that here, the image is acquired as pixels by an array of sensors. For pixel $i$, the acquired pixel value by the corresponding sensor is $\tilde{x}_i$, which is different from the true pixel value $x_i$ due to the measurement noise. Using the same treatment as (5), the acquired pixel value $\tilde{x}_i$ is given by

$$\tilde{x}_i = \hat{x}_i + \delta_i,$$
$$E(\hat{x}_i) = \mathrm{var}(\hat{x}_i) = x_i, \qquad i = 1,...,N, \qquad (10)$$
$$E(\delta_i) = 0, \ \mathrm{var}(\delta_i) = \rho^2,$$

where $\hat{x}_i$ are independent random variables with Poisson distribution, and $\delta_i$ are independent and identically distributed random variables. In (10), we allow the additive noise to have a different power than that in (4) because the sensors may have different operating dynamic ranges in different architecture.

*B. Comparison with pinhole aperture imaging (PAI)*

The LCI is closely related to the PAI as illustrated in Figure 2. If the sensing matrix used in the LCI is the identity matrix, i.e., if each measurement from the sensor of Figure 1 is made when only one of the aperture elements is open and all others are closed, then a measurement in LCI is equivalent to a pixel value in the PAI when the pinhole is placed at the location of sensor, see Figure 2. That is, if the sensor in the LCI is the same as the sensors in the PAI, and if the identity matrix is used as the sensing matrix in LCI, then LCI and PAI have the same SNR. However, the result of Proposition 1 is obtained because the modified Hadamard matrix is used as sensing matrix instead of the identity matrix. The modified Hadamard matrix provides an SNR gain in LCI. The following result is obtained by computing the variance of $\tilde{x}_i$ from (10).

**Proposition 2.**

The SNR of the image $\tilde{x}$ in PAI is given by

$$\mathrm{SNR}^{PAI} = X^0 / \sqrt{X^0 + N\rho^2}, \qquad (11)$$

where $X^0$ is the total signal power given in (7), and $\rho^2$ is the variance of the additive noise given in (10). Furthermore, the following estimate holds

$$\frac{\mathrm{SNR}^{LCI}}{\mathrm{SNR}^{PAI}} \geq \frac{\sqrt{X^0 + N\rho^2}}{\sqrt{2X^0 + 4\sigma^2}} \approx \frac{1}{\sqrt{2}} \sqrt{1 + \left(\frac{\sqrt{N}\rho}{\sqrt{X^0}}\right)^2}. \qquad (12)$$

An important observation from Proposition 2 is that the total SNR in PAI, $\mathrm{SNR}^{PAI}$, is not only a function of $X^0$ and $\rho^2$, like Proposition 1, but also a function of the image resolution $N$, unlike Proposition 1. The significance of the Proposition 2 is that the SNR of PAI decreases as the image resolution increases.

It is more revealing if we consider the ratio of the SNRs for LCI and PAI, as given in (12), which shows that the SNR of the LCI is higher than that of the PAI by an order of $\sqrt{N}$. One corollary is that no matter what sensors and quantization levels are used in two architectures (which determine the relative sizes of $\sigma$ and $\rho$), the LCI will always outperform the PAI if the image resolution is high enough, i.e., if $N$ is large enough.

For any given resolution $N$, we can make the following remarks. First, as shown in (12), in the worst case scenario, the SNR of LCI can only be lower than that of PAI by a factor of $\sqrt{2}$, which is about 1.5 dB. That is, the SNR of LCI can be no more than 1.5dB worse than that of PAI under any circumstance.

Secondly, the SNR of LCI is much better than PAI if the total shot noise is low, which happens if the scene is faint. This shows that LCI can have much better performance in low lighting environment, such as in surveillance or astronomy. In other words, LCI outperforms PAI in cases where SNR is concerned the most, which is when the shot noise is low. When the shot noise is high, the SNR is high also, causing very little concern about it.

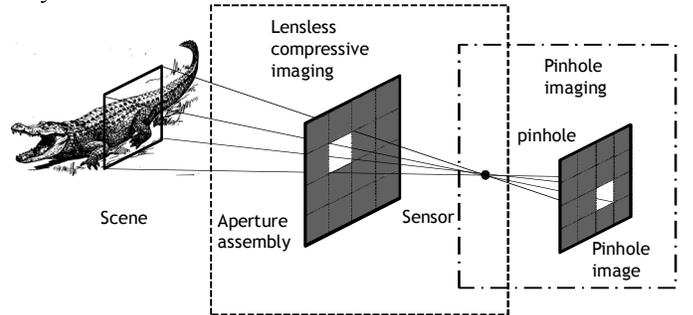

**Figure 2.** Relationship between LCI and PAI.

*C. Comparison with lens aperture imaging (LAI)*

To compare the LCI with the LAI, we assume that the sensor in the LCI of Figure 1 has a finite, nonzero, size. A nonzero size sensor in LCI will introduce blurring into the image, but the blurring can be removed or reduced during reconstruction [13]. It is out of scope of this paper to consider the blurring or how to reduce the blurring. We will simply consider the blurred image to be the desired image that we want to acquire, and there is no loss of rigor in doing so. This is because in the LAI, a perfectly non-blurred image is obtained only when the image plane, i.e., the plane of sensors, is placed exactly at the focal plane of the lens. In reality, this would never be possible, because just like we can never make an infinitesimal sensor in realty, we can never place an image plane at the exactly location of the focal plane in reality. Therefore, in LAI, there is always a blurring in the image due to the imaging plane not exactly being at the focal plane, even if we assume that the lens itself is perfectly made, which is also never be possible in realty.

Therefore, in this subsection, when comparing with LAI, we assume the sensor in LCI has a nonzero size, and we compare it with an LAI in which the image plane is not placed at the focal plane so that the images in both architectures have exactly the same amount of blurring. This is illustrated in Figure 3.

As shown in Figure 3 (a), we assume that the area of sensor in LCI is given by $S_{sensor}$. The area of the lens in LAI is given by $S_{lens}$ as shown in Figure 3 (b). The areas $S_{sensor}$ and $S_{lens}$ may be very different, but for the two architectures to have the same amount of blurring, the point spread functions, as illustrated in Figure 3, are assumed to be the same. In other words, we assume the imaging plane of LAI is placed appropriately away from the focal plane so that the point spread function matches that of LCI due to non-zero size sensor.

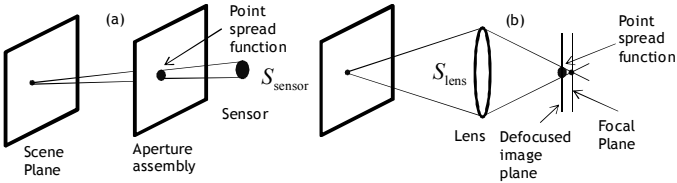

**Figure 3.** Blurred images in LCI and LAI. (a) Blurring due to nonzero size sensor in LCI. (b) Blurring due to displacement of image plane away from focal plane in LAI.

When comparing with PAI, we also assume that the size of the pinhole is nonzero, and the area of the hole is same as the area of the sensor in LCI, which is $S_{sensor}$. Note that $S_{sensor}$ is area of the sensor in LCI, or the area of pinhole in PAI, it is not the area of the sensors in PAI or LAI.

We can define the gain of lens as
$$g = S_{lens}/S_{sensor}. \quad (13)$$
In comparing the images in LAI and PAI, we find that they are the same, with the exception that the scene appears $g$ times brighter in LAI due to the gain of lens. This is because the amount of light rays arriving at the image plane when a lens is used is $g$ times more than that when a pinhole is used since the area of the lens is $g$ times larger than the pinhole. Consequently, a scene which is found to have the brightness of $X^0$ in PAI will be found to have the brightness of $gX^0$ in LAI. Therefore, the following Proposition follows directly from Proposition 2 by replacing $X^0$ by $gX^0$.

*Proposition 3.*

The SNR of the image $\tilde{x}$ in LAI is given by
$$\mathrm{SNR}^{LAI} = gX^0 \Big/ \sqrt{gX^0 + N\rho^2}. \quad (14)$$

Further, the following estimate holds
$$\frac{\mathrm{SNR}^{LCI}}{\mathrm{SNR}^{LAI}} \geq \frac{\sqrt{gX^0 + N\rho^2}}{g\sqrt{2X^0 + 4\sigma^2}} \approx \frac{1}{\sqrt{2g}}\sqrt{1+\left(\frac{\sqrt{N}\rho}{\sqrt{gX^0}}\right)^2}. \quad (15)$$

Comparing Propositions 2 and 3, we find that the SNR in LAI is higher than that in PAI because of the gain of the lens, $g$. Despite having a higher value, the SNR of LAI exhibits a same characteristic as that of PAI, namely, the SNR decreases as the image resolution $N$ increases.

Equation (15) shows that for given configurations of the two architectures, the LCI can have a higher SNR than the LAI if the image resolution is high enough, or if the scene is dim enough. Specifically, LCI will have a higher SNR if the total additive noise $\sqrt{N}\rho$ in LAI, which is an increasing function of the image resolution $N$, is higher than the total shot noise $\sqrt{gX^0}$ in LAI, which is a constant independent of image resolution.

## IV. SIMULATION

We demonstrate the behavior of the SNRs of LCI and PAI, as a function of image resolution $N$, by a simulation. We assume a scene has a fixed brightness $X^0$, in terms of number of photons. Then these photons are randomly assigned to the $N$ pixels of the image $x$ with a uniform distribution, so that the total number of photons in the image is $X^0$. The Poisson distribution is used to create shot noise, and the Gaussian distribution of variance $\sigma^2 = \rho^2$ is used to create additive noise. In LCI, the noise is added to the measurements, and the reconstructed image $\tilde{x}$ is obtained from the contaminated measurements by inverting the sensing matrix $A$ of (2). In PAI, the noise is added to the pixels of $x$ to obtain the contaminated image $\tilde{x}$. We then compute the SNR of the image $\tilde{x}$ for different values of image resolution $N$, and plot the results, together with the values obtained from theoretical analysis of previous section. The results are presented in Figure 4, in which the following parameters are used
$$X^0 = 10^7, \sigma = \rho = 5. \quad (16)$$

It can be observed from Figure 4 that the SNR in LCI is a constant with respect to the image resolution $N$, while SNR in PAI decreases as the image resolution increases. Furthermore, the simulation results match very well with the theoretical analysis of the previous sections.

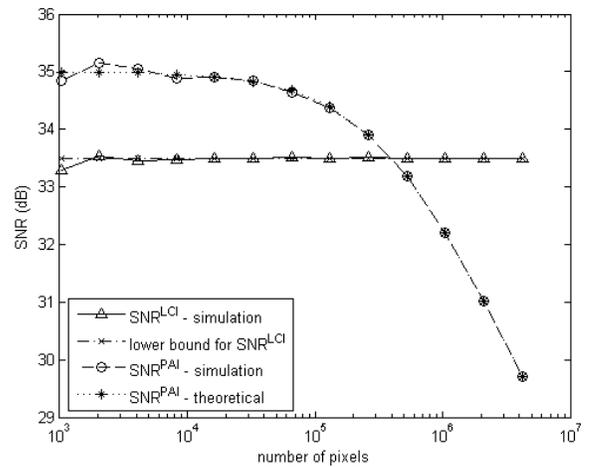

**Figure 4.** Signal to noise ratio of LCI and PAI as functions of image resolution (number of pixels). The lower bound for $\mathrm{SNR}^{LCI}$ is given by (9) and the theoretical result for $\mathrm{SNR}^{PAI}$ is given by (11).